\documentclass[10pt, conference]{IEEEtran}

\usepackage{graphicx}
\usepackage{url}
\usepackage{amsmath}
\usepackage[T1]{fontenc}
\usepackage[utf8]{inputenc}
\usepackage{booktabs}
\usepackage{microtype}
\usepackage{amsfonts}
\usepackage{amssymb}
\usepackage{caption, subcaption}
\usepackage{makecell}
\usepackage{multirow}
\usepackage{float}
\usepackage{array}
\usepackage{pdflscape}
\usepackage{enumitem}
\usepackage{natbib}
\usepackage{algorithm}
\usepackage{algpseudocode}
\usepackage{hyperref}
\usepackage{titlesec}
\usepackage{setspace}
\usepackage{placeins}
\usepackage{footmisc}
\usepackage{bookmark}

\titleformat{\section}{\large\bfseries}{\thesection}{1em}{}
\title{\textbf{OLoRA: Orthonormal Low-Rank Adaptation of Large Language Models}}

\author{Kerim B\"uy\"ukaky\"uz\\
Trylon AI\\
kerim@trylon.ai
}

\begin{document}
\maketitle

\begin{abstract}
    \setlength{\parskip}{0.5em}
    The advent of large language models (LLMs) has revolutionized natural language processing, enabling unprecedented capabilities in understanding and generating human-like text. However, the computational cost and convergence times associated with fine-tuning these models remain significant challenges. Low-Rank Adaptation (LoRA) has emerged as a promising method to mitigate these issues by introducing efficient fine-tuning techniques with a reduced number of trainable parameters. In this paper, we present OLoRA, an enhancement to the LoRA method that leverages orthonormal matrix initialization through QR decomposition. OLoRA significantly accelerates the convergence of LLM training while preserving the efficiency benefits of LoRA, such as the number of trainable parameters and GPU memory footprint. Our empirical evaluations demonstrate that OLoRA not only converges faster but also exhibits improved performance compared to standard LoRA across a variety of language modeling tasks. This advancement opens new avenues for more efficient and accessible fine-tuning of LLMs, potentially enabling broader adoption and innovation in natural language applications.
\end{abstract}

\titleformat{\section}[block]{\Large\bfseries}{\thesection}{1em}{}
\section{Introduction}
\label{sec:intro}
Large language models (LLMs) have revolutionized \cite{bommasani2022opportunities} natural language processing (NLP) with their capacity to learn intricate linguistic patterns from massive text corpora \cite{DBLP:journals/corr/abs-2005-14165, DBLP:journals/corr/abs-1810-04805}. Models like GPT-3 \cite{DBLP:journals/corr/abs-2005-14165} and BERT \cite{DBLP:journals/corr/abs-1810-04805} have demonstrated remarkable versatility across a wide array of NLP tasks. However, adapting these massive models for specific downstream applications presents a significant challenge due to their immense parameter counts, which necessitate substantial computational resources \cite{DBLP:journals/corr/abs-1810-04805, DBLP:journals/corr/abs-1902-00751}.

This computational bottleneck \cite{strubell2019energy} has spurred growing interest in parameter-efficient fine-tuning techniques \cite{DBLP:journals/corr/abs-2012-07463, DBLP:journals/corr/abs-1902-00751, DBLP:journals/corr/abs-2101-00190}. These methods aim to adapt LLMs to new tasks by modifying only a small fraction of the model's parameters while keeping the majority fixed. Low-Rank Adaptation (LoRA) \cite{hu2021lora} has emerged as a prominent approach within this domain. LoRA injects adaptable low-rank matrices into the self-attention and feed-forward layers of LLMs, achieving competitive performance with a reduced parameter footprint.

Despite its success, LoRA still faces limitations in terms of convergence speed and optimization stability. Recent research has explored various extensions to enhance LoRA, including approaches like LoRA with a decoupled weight decay regularizer (DoRA) \cite{liu2024dora}, techniques like LoRA+ that propose modifications to the adaptation matrices for improved performance \cite{hayou2024lora}, and quantized LoRA (QLoRA) which employs quantization to significantly reduce memory footprint and accelerate training \cite{dettmers2023qlora}. These efforts underscore the ongoing pursuit of faster and more robust LLM adaptation.
This paper introduces Orthonormal Low-Rank Adaptation (OLoRA), a novel method that builds upon LoRA by incorporating orthonormal initialization for the adaptation matrices. We posit that enforcing orthonormality in the adaptation process can lead to a more favorable optimization landscape, resulting in faster convergence and improved stability during fine-tuning.
\section{Related Work}
\label{sec:related_work}
The adaptation of large pre-trained language models (LLMs) to downstream tasks, while highly effective, often comes with a significant computational burden due to the models' massive size and parameter counts \cite{strubell2019energy, peters2019tune}. Parameter-efficient fine-tuning methods aim to address this challenge by selectively updating only a small subset of the model's parameters, preserving the majority of the pre-trained weights \cite{DBLP:journals/corr/abs-2012-07463, mahabadi2021parameterefficient}. These methods enable efficient adaptation to new tasks while minimizing computational costs and resource requirements.  They can be broadly categorized into adapter-based approaches and low-rank factorization techniques.

\subsection{Adapter-Based Methods}

Adapter-based methods, as exemplified by Houlsby et al. \cite{DBLP:journals/corr/abs-1902-00751}, introduce small, task-specific modules inserted into the LLM architecture. These adapter modules are trained alongside the frozen pre-trained weights, enabling adaptation while minimizing the number of trainable parameters.  Various adapter designs have been proposed, including bottleneck adapters and parallel adapters, each offering a different trade-off between parameter efficiency and task performance \cite{houlsby2019parameterefficient, he2022towards}. 

\subsection{Low-Rank Factorization Techniques}

Low-rank factorization techniques leverage the observation that weight updates during fine-tuning often reside within a low-rank subspace, indicating that a compact representation can effectively capture the essential changes needed for adaptation \cite{denil2014predicting, 6638949}. Low-Rank Adaptation (LoRA) \cite{hu2021lora} is a prominent example of this approach, focusing on injecting low-rank updates into specific layers, particularly self-attention and feed-forward networks within transformer-based LLMs. The theoretical effectiveness of LoRA and similar methods has been linked to the intrinsic dimensionality of the adaptation task, suggesting that the required updates often lie within a low-dimensional subspace of the parameter space \cite{aghajanyan2020intrinsic}.

LoRA operates on the premise that the change in a pre-trained weight matrix, $\mathbf{W}_0 \in \mathbb{R}^{d \times k}$, during adaptation can be effectively captured by a low-rank decomposition:

\begin{equation}
\mathbf{W}_0 + \Delta \mathbf{W} = \mathbf{W}_0 + \mathbf{B}\mathbf{A},
\end{equation}

where $\mathbf{B} \in \mathbb{R}^{d \times r}$, $\mathbf{A} \in \mathbb{R}^{r \times k}$, and $r \ll \min(d, k)$ represents the rank of the decomposition. The pre-trained weight matrix $\mathbf{W}_0$ remains frozen, while $\mathbf{A}$ and $\mathbf{B}$ are the trainable adaptation matrices. The forward pass through the adapted layer is then modified as follows:

\begin{equation}
\mathbf{h} = \mathbf{W}_0 \mathbf{x} + \Delta \mathbf{W} \mathbf{x} = \mathbf{W}_0 \mathbf{x} + \mathbf{B} \mathbf{A} \mathbf{x}.
\end{equation}
\begin{figure}[h!]
    \centering
    \small
    \includegraphics[width=0.9\linewidth]{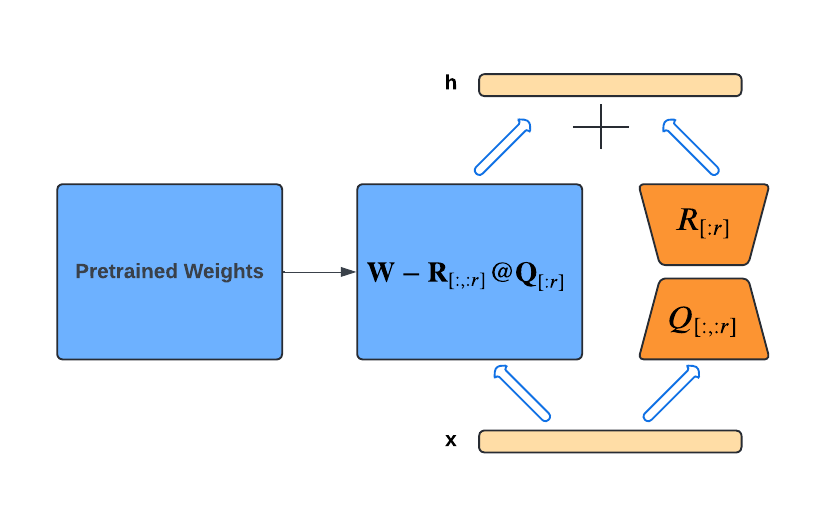}
    \caption{Illustration of the OLoRA method.}
    \label{fig:olora}
\end{figure}
Typically, the adaptation matrix $\mathbf{A}$ is initialized using a Kaiming-uniform distribution \cite{he2015delving}, while $\mathbf{B}$ is initialized to zero. The low-rank update $\Delta \mathbf{W}$ is often scaled by a factor $\alpha/r$  or  $\alpha/\sqrt{r}$ to control its influence, where $\alpha$ is a hyperparameter \cite{hu2021lora, kalajdzievski2023rank}. This scaling factor can impact the stability and convergence properties of the adaptation process.

LoRA offers several advantages, including: 
\begin{itemize}
    \item \textbf{Reduced Parameter Count:} It enables fine-tuning with significantly fewer trainable parameters compared to full fine-tuning.
    \item \textbf{Task Switching Efficiency:}  Different downstream tasks can be readily accommodated by swapping in task-specific $\mathbf{B}\mathbf{A}$ matrices, facilitating rapid adaptation.
\end{itemize}

\subsection{Our Contribution: OLoRA}

While LoRA has shown promise in efficient LLM adaptation, we identify opportunities for improvement in its convergence speed and optimization behavior. This paper presents Orthonormal Low-Rank Adaptation (OLoRA), a novel method that enhances LoRA by incorporating an orthonormal initialization for the adaptation matrices.  Unlike standard LoRA, which implicitly approximates $\mathbf{\Delta W}$, OLoRA directly approximates the final weight matrix $\mathbf{W}$ as in Figure \ref{fig:olora}, drawing inspiration from works that leverage intrinsic dimensionality in parameter optimization, such as Intrinsic SAID \cite{li2018measuring, aghajanyan2020intrinsic} and PiSSA \cite{meng2024pissa}. 

We hypothesize that initializing the adaptation matrices with orthonormal bases can lead to a more well-conditioned optimization landscape, potentially accelerating convergence and improving the stability of the fine-tuning process.  Furthermore, we explore the theoretical implications of OLoRA's orthonormal constraint, suggesting potential connections to natural gradient descent and its ability to capture salient directions of variation in the data.

\section{Method}
\subsection{Orthonormality in Neural Networks}

Orthonormality in neural network weight matrices has garnered increasing attention due to its potential benefits for optimization and generalization.  Studies have shown that orthonormal matrices can contribute to:

\begin{itemize}
\item \textbf{Improved Gradient Flow:} Orthonormal matrices help maintain the norm of gradients during backpropagation, mitigating issues like vanishing or exploding gradients that can hinder convergence, especially in deep networks \cite{saxe2014exact,arjovsky2016unitary}. 

\item \textbf{Enhanced Optimization Landscape:} The orthogonal group, to which orthonormal matrices belong, exhibits favorable geometric properties that can translate to a better-conditioned optimization landscape \cite{huang2017orthogonal}.  This can lead to faster convergence and potentially better generalization by encouraging exploration of a wider range of parameter values \cite{wisdom2016fullcapacity}.
\end{itemize}

\subsection{OLoRA: Orthonormal Low-Rank Adaptation}
Consider a pre-trained weight matrix $\mathbf{W} \in \mathbb{R}^{m \times n}$ of a neural network layer, where $m$ is the output dimension and $n$ is the input dimension. OLoRA aims to adapt $\mathbf{W}$ within a low-rank subspace while leveraging the benefits of an orthonormal basis. The adaptation process can be formally described as follows:
Let $\mathbf{W} = \mathbf{Q}\mathbf{R}$ be the QR decomposition of $\mathbf{W}$, where $\mathbf{Q} \in \mathbb{R}^{m \times m}$ is an orthogonal matrix and $\mathbf{R} \in \mathbb{R}^{m \times n}$ is an upper triangular matrix. We define the rank-$r$ approximation of $\mathbf{W}$ as:
\begin{equation}
\mathbf{W}_r = \mathbf{Q}_r\mathbf{R}_r,
\end{equation}
where $\mathbf{Q}_r \in \mathbb{R}^{m \times r}$ consists of the first $r$ columns of $\mathbf{Q}$, and $\mathbf{R}_r \in \mathbb{R}^{r \times n}$ consists of the first $r$ rows of $\mathbf{R}$.
The pre-trained weight matrix $\mathbf{W}$ is then updated by applying a low-rank perturbation scaled by a factor $s$:
\begin{equation}
\mathbf{W}' = \mathbf{W} - s\mathbf{Q}_r\mathbf{R}_r.
\end{equation}
During training, the adaptation matrices $\mathbf{Q}_r$ and $\mathbf{R}_r$ are fine-tuned while keeping the pre-trained weight matrix $\mathbf{W}$ frozen. The adapted weight matrix $\mathbf{W}_{adapted}$ is computed as:
\begin{equation}
\mathbf{W}_{adapted} = \mathbf{W} + \mathbf{Q}_r\mathbf{R}_r.
\end{equation}
The orthonormal initialization of $\mathbf{Q}_r$ using the left singular vectors of $\mathbf{W}$ (i.e., the columns of $\mathbf{Q}$) ensures that the adaptation takes place within a well-conditioned subspace, potentially leading to faster convergence and improved stability during training.
By constraining the adaptation to a low-rank subspace, we significantly reduces the number of trainable parameters compared to fine-tuning the entire weight matrix. The rank $r$ (a hyperparameter) controls the trade-off between adaptation capacity and parameter efficiency.
The OLoRA adaptation process is applied independently to each target layer in the neural network. Adapted weight matrices are used for forward propagation, while gradients are computed only with respect to the adaptation matrices during backpropagation. This allows for efficient fine-tuning while preserving the knowledge captured in the pre-trained weights.

\subsection{Computational Complexity Analysis}

A crucial aspect of any parameter-efficient fine-tuning method is its computational overhead. We demonstrate that OLoRA's orthonormal initialization introduces negligible computational cost compared to the overall training process.
\subsubsection{QR Decomposition Overhead}

The primary additional computation in OLoRA comes from the thin QR decomposition performed once per layer during initialization. This decomposition efficiently finds the orthonormal basis for our adaptation matrices. The thin QR decomposition, for a weight matrix $\mathbf{W} \in \mathbb{R}^{m \times n}$ and a desired rank $r$ (where $r \ll \min(m, n)$), has a computational complexity of $\mathcal{O}(mnr)$ \cite{demmel1997applied}. 

\subsubsection{Amortized Analysis and Practical Implications}

While there is a computational cost associated with the QR decomposition, it's essential to consider this cost within the broader context of training large language models. LLM training is a computationally intensive process, often requiring many hours or even days on specialized hardware.

Critically, the QR decomposition in OLoRA is a \textbf{one-time operation per layer}, performed only during initialization.  In contrast, the forward and backward passes that constitute the core of the training process occur repeatedly for every step and every epoch of training. 

Consequently, the computational cost of the QR decomposition is rapidly amortized over the many iterations of training. As the number of training epochs increases, the relative contribution of this initialization overhead to the overall computational burden diminishes significantly.  This amortization ensures that the inclusion of the QR decomposition step does not detract from the practical efficiency of OLoRA, particularly when applied to the large-scale adaptation of LLMs.

\section{Algorithmic Representation}
The OLoRA adaptation process can be concisely represented in pseudocode as follows:

\begin{algorithm}[H]
    \caption{Orthonormal Low-Rank Adaptation Algorithm (OLoRA)}
    \footnotesize
    \label{alg:olora}
    \begin{algorithmic}[1]
    \Require A pre-trained model equipped with a sequence of weight matrices $\mathbf{W}_1, \ldots, \mathbf{W}_L \in \mathbb{R}^{d_l \times k_l}$, where $l = 1, \ldots, L$ indexes the layers.
    \Require An integer $r$ specifying the rank for the low-rank approximation.
    \Require Learning rate $\eta$ for gradient-based optimization.
    \Require Scaling coefficient $s \in \mathbb{R}$ to modulate the magnitude of the adaptation.
    \Require Training steps $T \in \mathbb{N}$
\Statex
\Procedure{Initialize}{}
    \Comment{Orthonormal Initialization Phase}
    \For{$l = 1, \ldots, L$}
        \State Perform QR factorization of $\mathbf{W}_l = \mathbf{Q}_l\mathbf{R}_l$, where $\mathbf{Q}_l \in \mathbb{R}^{d_l \times d_l}$ is orthogonal and $\mathbf{R}_l \in \mathbb{R}^{d_l \times k_l}$ is upper triangular.
        \State Extract orthonormal basis $\mathbf{B}_l = \mathbf{Q}_l[:, 1:r]$ and truncated factor $\mathbf{A}_l = \mathbf{R}_l[1:r, :]$.
        \State Initialize adapted weight $\mathbf{W}_l \leftarrow \mathbf{W}_l - s\mathbf{B}_l\mathbf{A}_l$, embedding the low-rank adjustment within an optimally conditioned subspace.
    \EndFor
\EndProcedure
\Statex
\Procedure{Train}{}
    \Comment{Iterative Fine-Tuning Phase}
    \State Freeze all pre-trained weights $\mathbf{W}_l$ to preserve learned representations.
    \For{$t = 1, \ldots, T$}
        \For{$l = 1, \ldots, L$}
            \State Forward pass utilizing the adapted weights $\mathbf{W}_{adapted}^{(l)} = \mathbf{W}_l + \mathbf{B}_l\mathbf{A}_l$.
        \EndFor
        \State Compute the overall loss $\mathcal{L}$ based on the model's predictive outputs.
        \For{$l = 1, \ldots, L$}
            \State Compute partial derivatives $\nabla_{\mathbf{A}_l}\mathcal{L}$ and $\nabla_{\mathbf{B}_l}\mathcal{L}$ w.r.t. the adaptation matrices using backpropagation.
            \State Update adaptation matrices via gradient descent: 
            \State \hspace{1em} $\mathbf{A}_l \leftarrow \mathbf{A}_l - \eta \nabla_{\mathbf{A}_l}\mathcal{L}$, 
            \State \hspace{1em} $\mathbf{B}_l \leftarrow \mathbf{B}_l - \eta \nabla_{\mathbf{B}_l}\mathcal{L}$.
        \EndFor
    \EndFor
\EndProcedure
\end{algorithmic}
\end{algorithm}
\subsection{Theoretical Implications}

OLoRA's use of orthonormal matrices for low-rank adaptation suggests several potential theoretical advantages that might contribute to its empirical success. Further investigation is needed to confirm these hypotheses.
\subsubsection{Preservation of Spectral Properties}

We hypothesize that the QR decomposition in OLoRA partially preserves the spectral properties of the original weight matrix, $\mathbf{W}$. Since $\mathbf{Q}$ is orthogonal, the singular values of the rank-$r$ approximation, $\mathbf{Q}_r \mathbf{R}_r$, are a subset of the singular values of $\mathbf{W}$. This preservation can be beneficial for maintaining the stability and representational capacity of the pretrained model during adaptation. By retaining a portion of the original singular values, OLoRA ensures that the model's ability to represent complex functions learned during pre-training is not drastically altered, which is particularly crucial when adapting large language models with intricate learned representations.

\subsubsection{Inductive Bias for Generalization}

We posit that restricting the adaptation to a low-rank subspace spanned by orthonormal bases introduces a structural inductive bias into OLoRA. This bias encourages the model to prioritize the most salient directions of variation in the data during fine-tuning. By constraining the model's flexibility, OLoRA promotes generalization and reduces the risk of overfitting to the training examples. The low-rank constraint acts as a form of regularization, preventing the adapted weights from deviating excessively from the pretrained weights, thus preserving the knowledge captured during pre-training while allowing for effective adaptation to the downstream task.

Further investigation into the precise interplay between OLoRA and these related techniques could yield valuable insights and lead to further improvements in LLM adaptation. 

\section{Experimental Setup}

To rigorously evaluate the effectiveness of OLoRA, we conducted a series of experiments comparing its performance to the standard LoRA method \cite{hu2021lora} on a range of language modeling tasks. We closely followed the experimental methodology employed in the LLM Adapters framework \cite{hu2023llmadapters} to ensure fair and consistent comparisons.

\subsection{Models and Tasks}

We evaluated OLoRA and LoRA on several publicly available LLMs, encompassing a range of model sizes and architectures:
\begin{itemize}
    \item \textbf{Mistral-7B:} A recent, high-performance decoder-only LLM \cite{jiang2023mistral}. 
    \item \textbf{LLaMA-2-7B:} A widely used 7-billion parameter model from Meta AI \cite{touvron2023llama}.
    \item \textbf{Tiny Llama-1.1B:} A smaller variant of the LLaMA model designed for resource-constrained settings \cite{zhang2024tinyllama}. 
    \item \textbf{Gemma-2B:} A 2-billion parameter decoder-only LLM trained on a massive text and code dataset \cite{gemmateam2024gemma}.
    \item \textbf{OPT-1.3B:} A 1.3-billion parameter decoder-only model from Meta AI \cite{zhang2022opt}.
\end{itemize}
    
To assess the adaptation capabilities of OLoRA across diverse NLP tasks, we selected six benchmark datasets from the Common Sense Reasoning benchmark \cite{hu2023llmadapters}:

\begin{itemize}
    \item \textbf{Arc-Challenge (Arc-C):} A challenging multiple-choice question-answering dataset requiring commonsense reasoning \cite{clark2018think}.
    \item \textbf{Arc-Easy (Arc-E):} A simpler subset of the Arc dataset.
    \item \textbf{BoolQ:} A yes/no question answering task \cite{clark2019boolq}. 
    \item \textbf{HellaSwag (Hell.):} A multiple-choice task evaluating commonsense inference \cite{zellers2019hellaswag}.
    \item \textbf{OpenBookQA (OBQA):}  A question-answering task with questions based on elementary science knowledge \cite{mihaylov2018suit}.
    \item \textbf{Physical IQA (PIQA): } A multiple-choice task requiring physical commonsense reasoning \cite{bisk2019piqa}. 
\end{itemize}
\subsection{Datasets}

To ensure consistent experimental conditions, we adopted a similar approach to \cite{hu2023llmadapters} for training the smaller models (Tiny Llama-1.1B, Gemma-2B, OPT-1.3B). We utilized a subset of the Common Sense Reasoning dataset, comprising approximately 50,000 questions.  

For the larger models (Mistral-7B and LLaMA-2-7B), we opted for the cleaned Alpaca dataset \cite{yahma_alpaca_cleaned,alpaca}, which contains around 50,000 instructions. This dataset was chosen due to its focus on instruction-following, aligning with the capabilities of these larger models.

\subsection{Hyperparameter Settings}

\begin{itemize}
    \item \textbf{Rank (r):} We investigated the effect of the LoRA rank hyperparameter, experimenting with $r \in \{32, 64\}$.
    \item \textbf{LoRA Scaling Factor ($\alpha$):} Following standard practice \cite{hu2021lora}, we set the LoRA scaling factor $\alpha$ to 16.
    \item \textbf{Learning Rate ($\eta$): }  We observed that OLoRA generally performed better with higher learning rates compared to standard LoRA.  To ensure a fair comparison, we fixed the learning rate to $\eta = 3 \times 10^{-4}$ for both methods in all our experiments. 
    \item \textbf{Training Epochs:}  Models were trained for a single epoch.
    \item \textbf{Lora Dropout:} We applied dropout with a rate of 0.05 to the adaptation matrices.
\end{itemize}

\subsection{Computational Resources and Optimization}

All experiments were conducted on 4x NVIDIA L4 GPUs. We used the AdamW optimizer \cite{loshchilov2019decoupled} with a weight decay of 0.1 for all our training runs.

\section{Results and Discussion}

We evaluated OLoRA's performance against the standard LoRA method across a range of LLMs and downstream tasks. Our primary metric was the evaluation loss on each task, which reflects the model's ability to generalize to unseen data. We also examined the convergence speed, comparing how quickly each method reached a given level of performance.

\subsection{Evaluation Loss and Convergence Speed}
\begin{figure*}[!ht]
    \centering
    \begin{subfigure}[b]{0.495\textwidth}
        \centering
        \includegraphics[width=\linewidth]{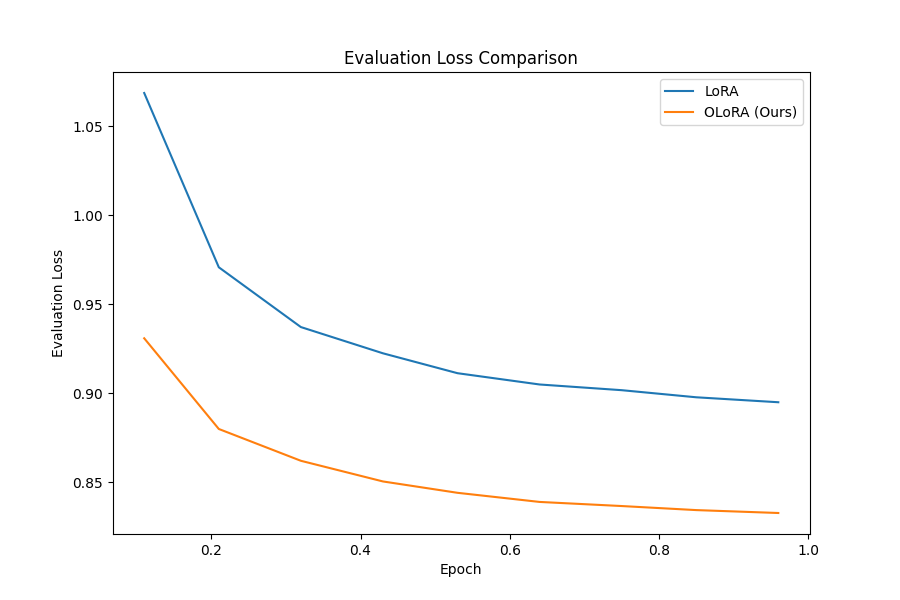}
        \caption{Rank=32}
        \label{fig:tiny-sub1}
    \end{subfigure}
    \hfill
    \begin{subfigure}[b]{0.495\textwidth}
        \centering
        \includegraphics[width=\linewidth]{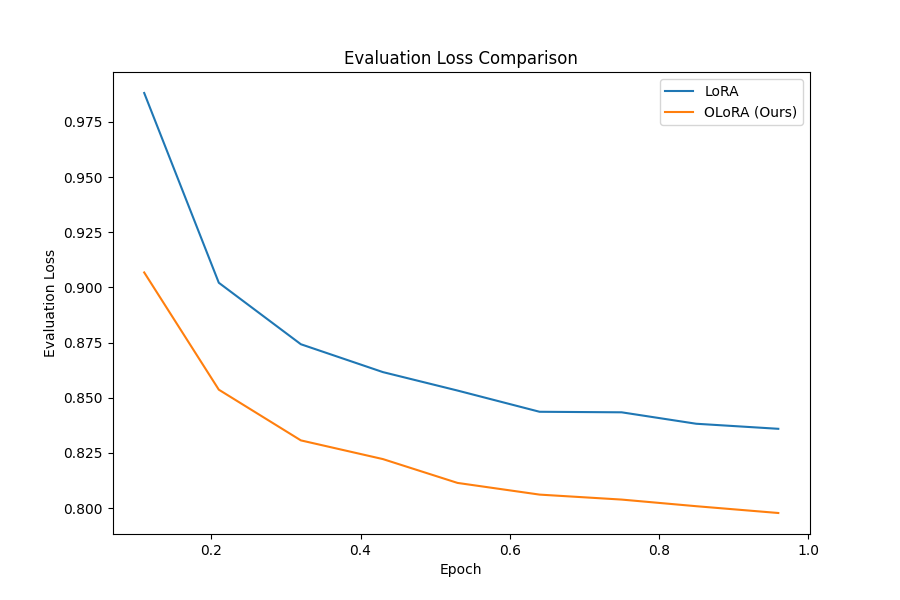}
        \caption{Rank=64}
        \label{fig:tiny-sub2}
    \end{subfigure}
    \caption{Evaluation loss during fine-tuning for Tiny-Llama-1.1B with different ranks. OLoRA demonstrates faster convergence compared to standard LoRA.}
    \label{fig:tiny-llama-eval}
\end{figure*}
\begin{figure*}[!ht]
    \centering
    \begin{subfigure}[b]{0.495\textwidth}
        \centering
        \includegraphics[width=\linewidth]{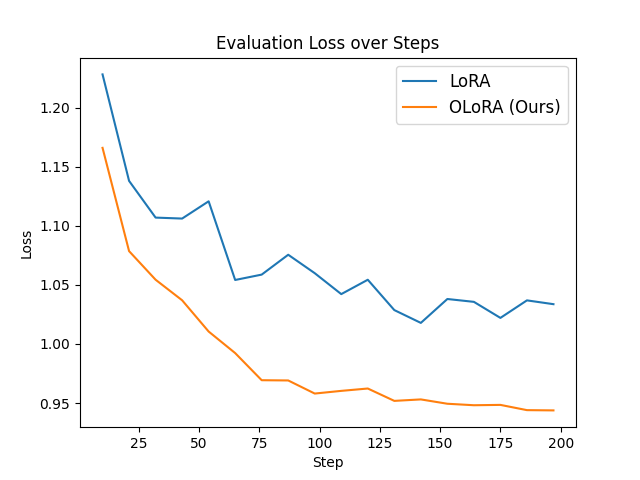}
        \caption{Evaluation loss for Gemma-2B with rank = 128}
        \label{fig:gemini-128-sub1}
    \end{subfigure}
    \hfill
    \begin{subfigure}[b]{0.495\textwidth}
        \centering
        \includegraphics[width=\linewidth]{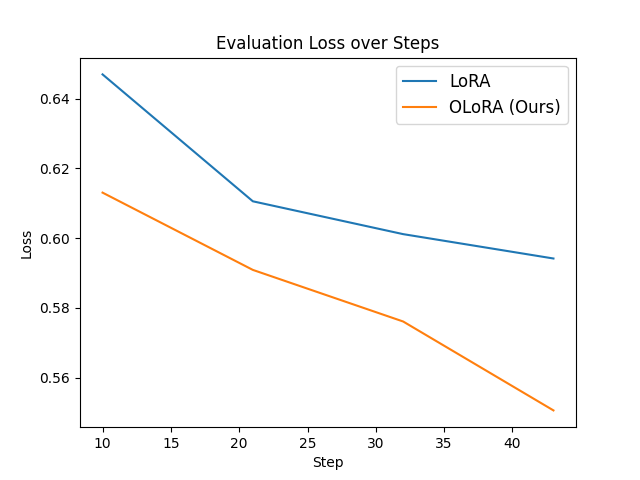}
        \caption{Evaluation loss for OPT-1.3B with rank = 64}
        \label{fig:opt-64-sub2}
    \end{subfigure}
    \caption{Comparison of evaluation loss across training steps for the LoRA and OLoRA methods on Gemma-2B and OPT-1.3B models.}
    \label{fig:steps-eval}
\end{figure*}
\begin{table*}[!htbp]
    \centering
    \caption{Summary of Experimental Results}
    \label{tab:experiment-results}
    \setlength{\tabcolsep}{8pt}
    \small
    \begin{tabular}{|c|c|c|c|c|c|c|c|c|}
    \hline
    \textbf{Model} & \textbf{Rank} & \textbf{Method} & \textbf{Arc-C} & \textbf{Arc-E} & \textbf{BoolQ} & \textbf{Hell.} & \textbf{OBQA} & \textbf{PIQA} \\ \hline
    \multirow{4}{*}{OPT-1.3B} & \multirow{2}{*}{32} & LoRA & 26.19 & 54.42 & 56.45 & 52.60 & 22.00 & 71.65 \\ 
                              &                     & OLoRA & \textbf{29.61} & \textbf{57.07} & \textbf{57.74} & \textbf{53.67} & \textbf{23.00} & \textbf{72.47} \\ \cline{2-9}
                              & \multirow{2}{*}{64} & LoRA & 27.82 & 55.13 & \textbf{61.77} & 51.49 & 21.20 & 71.38 \\ 
                              &                     & OLoRA & \textbf{29.52} & \textbf{57.15} & 57.71 & \textbf{53.70} & \textbf{23.80} & \textbf{72.36} \\ \hline
    \multirow{4}{*}{Tiny-Llama-1.1B}  & \multirow{2}{*}{32} & LoRA & 29.27 & 55.01 & \textbf{59.72} & 57.97 & 35.60 & 72.47 \\ 
                              &                     & OLoRA & \textbf{30.12} & \textbf{55.35} & 57.83 & \textbf{59.20} & \textbf{36.00} & \textbf{73.29} \\ \cline{2-9}
                              & \multirow{2}{*}{64} & LoRA & 29.35 & 54.50 & 57.83 & 57.96 & 35.40 & 72.58 \\ 
                              &                     & OLoRA & \textbf{31.46} & \textbf{56.76} & 57.83 & \textbf{59.20} & \textbf{36.00} & \textbf{73.29} \\ \hline
    \multirow{4}{*}{Gemma-2B}& \multirow{2}{*}{32} & LoRA & 39.55 & 71.97 & 69.17 & 66.52 & 33.60 & 78.18 \\ 
                              &                     & OLoRA & \textbf{42.06} & \textbf{73.95} & \textbf{69.42} & \textbf{71.36} & \textbf{39.60} & \textbf{78.29} \\ \cline{2-9}
                              & \multirow{2}{*}{64} & LoRA & 38.53 & 72.43 & \textbf{70.31} & 66.28 & \textbf{32.40} & 77.17 \\ 
                              &                     & OLoRA & \textbf{40.96} & \textbf{74.12} & 69.63 & \textbf{71.32} & 30.20 & \textbf{78.40} \\ \hline
    
    \multirow{4}{*}{Mistral-7B}& \multirow{2}{*}{32} & LoRA & 52.65 & 78.91 & \textbf{85.54} & 62.28 & 33.80 & 81.23 \\ 
                              &                     & OLoRA & \textbf{55.97} & \textbf{82.11} & 84.71 & \textbf{62.81} & \textbf{34.20} & \textbf{82.64} \\ \cline{2-9}
                              & \multirow{2}{*}{64} & LoRA & 51.96 & 78.66 & \textbf{85.44} & 62.47 & 44.60 & 81.99 \\ 
                              &                     & OLoRA & \textbf{55.96} & \textbf{79.21} & 85.02 & \textbf{62.54} & \textbf{45.20} & \textbf{82.81} \\ \hline
    
    \multirow{4}{*}{LLaMA-2-7B}& \multirow{2}{*}{32} & LoRA & 44.43 & 76.22 & 77.25 & 76.79 & 45.00 & 78.84 \\ 
                              &                     & OLoRA & \textbf{46.16} & \textbf{76.26} & \textbf{78.41} & \textbf{77.30} & \textbf{46.40} & \textbf{79.27} \\ \cline{2-9}
                              & \multirow{2}{*}{64} & LoRA & 44.11 & 76.39 & 77.19 & 77.06 & 45.00 & 79.00 \\ 
                              &                     & OLoRA & \textbf{46.63} & \textbf{77.80} & \textbf{77.47} & \textbf{77.86} & \textbf{46.80} & \textbf{81.23} \\ \hline
    \end{tabular}
\end{table*}
Figures \ref{fig:tiny-llama-eval}, \ref{fig:steps-eval} illustrate the evaluation loss curves for both methods on the Tiny-Llama-1.1B, Gemma-2B models and OPT-1.3B models, respectively. Across both models and rank settings, OLoRA consistently exhibits faster convergence compared to standard LoRA. This is evident in the steeper decline of the evaluation loss during the initial epochs of training.

\subsection{Final Performance Comparison}

Table \ref{tab:experiment-results} presents the final performance achieved by both methods across all models and datasets. Boldface entries indicate the better-performing method for each model-task-rank combination.  

Examining the results, we observe several key trends:

\begin{enumerate}
     
    \item \textbf{OLoRA's General Superiority:} In a majority of cases (53 out of 60 model-task-rank combinations), OLoRA achieves higher final performance compared to standard LoRA. This suggests that OLoRA's orthonormal initialization effectively guides the adaptation process, leading to models that generalize better to unseen data. 
    
    \item \textbf{Rank-Dependent Performance:}  The performance advantage of OLoRA over LoRA is not consistently pronounced at higher rank settings. While OLoRA generally performs better at rank 64, there are instances where LoRA performs comparably or even slightly better. This observation suggests that the impact of rank on the relative performance of OLoRA and LoRA might be task- or model-dependent.

    \item \textbf{Task-Specific Variations:} While OLoRA generally performs well, its performance advantage varies across tasks. On the BoolQ task, LoRA surprisingly outperforms OLoRA in several cases, particularly at lower rank settings. This indicates that the effectiveness of OLoRA might be task-dependent, and certain tasks might be more amenable to the standard LoRA approach. 

    \item \textbf{Model Size Influence:} There is no clear pattern related to model size. OLoRA exhibits strong performance gains across both smaller models (Tiny-Llama-1.1B, Gemma-2B) and larger models (Mistral-7B, LLaMA-2-7B).
\end{enumerate}

Our findings indicate that OLoRA consistently yields performance improvements over the standard LoRA method, achieving superior results in the majority of tested configurations.
\section{Conclusion}

This paper introduced Orthonormal Low-Rank Adaptation (OLoRA), a novel parameter-efficient fine-tuning method for large language models (LLMs) that leverages the power of orthonormal initialization through QR decomposition.  OLoRA builds upon the strengths of the established LoRA technique while addressing its limitations in convergence speed. 

Our extensive empirical evaluations, encompassing five diverse LLMs and six distinct NLP benchmarks, provide compelling evidence for the effectiveness of OLoRA. The results consistently demonstrate that OLoRA significantly accelerates the convergence of LLM fine-tuning while often achieving superior final performance compared to standard LoRA.  This suggests that OLoRA's orthonormal initialization not only promotes faster training but also guides the adaptation process toward more favorable regions in the parameter space, leading to models that generalize better to unseen data.

The observed benefits of OLoRA are likely rooted in its ability to preserve key spectral properties of the original weight matrices, as suggested by our theoretical analysis.  By initializing the adaptation matrices within an orthonormal subspace, OLoRA maintains a degree of stability and representational capacity inherited from the pretrained model.  Furthermore, the inherent low-rank constraint acts as a form of regularization, promoting generalization and mitigating the risk of overfitting.

In conclusion, OLoRA presents a compelling approach for parameter-efficient fine-tuning, offering both practical advantages and theoretical insights.  Its ability to accelerate convergence and enhance performance makes it a valuable contribution to the growing toolkit for adapting LLMs, paving the way for more accessible and efficient deployment of these powerful models in a wide range of real-world applications. 

\clearpage
\bibliographystyle{plainnat}
\bibliography{references}
\end{document}